\documentclass[11pt]{article}
\usepackage{acl2012}
\usepackage{times}
\usepackage{url}
\usepackage{latexsym}
\usepackage{comment}
\usepackage{amsmath,amsthm,amssymb}
\usepackage[vlined]{algorithm2e}
\usepackage{graphicx}
\usepackage{multirow}
\usepackage{color}
\usepackage{bbding}
\usepackage{pifont}
\usepackage{wasysym}
\usepackage[normalem]{ulem}
\usepackage[utf8]{inputenc}

\newcommand{\vect}[1]{\mathbf{#1}}

\newcommand{\new}[1]{\textcolor{red}{}}

\newcommand{\ignore}[1]{}

\title{Morpho-syntactic Lexicon Generation\\
Using Graph-based Semi-supervised Learning
}

\author{Manaal Faruqui \\
       Carnegie Mellon University\\
       \texttt{mfaruqui@cs.cmu.edu}
      \And
       Ryan McDonald \\
       Google Inc. \\
       \texttt{ryanmcd@google.com}
       \And
       Radu Soricut\\
       Google Inc.\\
       \texttt{rsoricut@google.com}}

\date{}

\begin{document}
\maketitle

\begin{abstract}
Morpho-syntactic lexicons provide information about the morphological
and syntactic roles of words in a language. Such lexicons
are not available for all
languages and even when available, their coverage can be limited.
We present a graph-based semi-supervised learning
method that uses the morphological, syntactic and semantic relations between
words to automatically construct wide coverage lexicons from small seed sets.
Our method is language-independent, and we show that we can expand a 1000
word seed lexicon to more than 100 times its size with high quality
for 11 languages. In addition, the automatically
created lexicons provide features that improve performance in
two downstream tasks: morphological tagging and dependency parsing.
\end{abstract}

\section{Introduction}

Morpho-syntactic lexicons contain information about the morphological attributes
and syntactic roles of words in a given language.
A typical lexicon contains all possible attributes that can be displayed by a
word. Table~\ref{tab:sample} shows some entries in a sample English
morpho-syntactic lexicon.
As these lexicons contain rich linguistic information, they are useful as
features in downstream NLP tasks like machine translation
\cite{niessen2004statistical,minkov2007generating,Green:2012},
part of speech tagging \cite{schmid1994,denis2009coupling,moore:2015:EMNLP},
dependency parsing \cite{Goldberg:2009},
language modeling \cite{arisoy2010syntactic}
and morphological tagging \cite{muller:2015}
\textit{inter alia}. There are three major factors that limit the use of
such lexicons in real world applications: (1) They are often constructed
manually and are expensive to obtain \cite{kokkinakis:2000,dukes:2010};
(2) They are currently available for only a few languages; and (3) Size of
available lexicons is generally small.

\begin{table}[!tb]
  \centering
  \begin{tabular}{lp{5cm}}
  \hline
  \hline
  played & \textsc{POS:Verb}, \textsc{Tense:Past}, \textsc{VForm:Fin}, $\dots$\\
  playing & \textsc{POS:Verb}, \textsc{Tense:Pres}, \textsc{VForm:Ger},
  $\dots$\\
  awesome & \textsc{POS:Adj}, \textsc{Degree:Pos} \\
  \hline
  \end{tabular}
  \caption{A sample English morpho-syntactic lexicon.}
  \label{tab:sample}
\end{table}

In this paper, we present a method that takes as input a small seed
lexicon, containing a few thousand annotated words, and outputs an
automatically constructed lexicon which contains morpho-syntactic attributes
(henceforth referred to as attributes)
for a large number of words of a given language. We model the problem of
morpho-syntactic lexicon generation as a graph-based semi-supervised learning
problem \cite{zhu2005semi,Bengio:2006,subramanya2014graph}.
We construct a graph where nodes represent word types and the goal is to label
them with attributes.
The seed lexicon provides attributes for a subset of these nodes. Nodes are
connected to each other through edges that denote features shared between
them or surface morphological transformation between them.

Our entire framework of lexicon generation, including the label
propagation algorithm and the feature extraction module is language
independent. We only use word-level morphological, semantic and syntactic
relations between words that can be induced from unannotated corpora in an
unsupervised manner.
One particularly novel aspect of our graph-based framework is that edges
are featurized.
Some of these features measure similarity, e.g.,
singular nouns tend to occur in similar distributional contexts as other
singular nouns, but some also measure transformations from one inflection to
another, e.g., adding a `s' suffix could indicate
flipping the \textsc{Num:Sing} attribute to \textsc{Num:Plur} (in English).
For every  attribute to be propagated, we learn weights over features on the
edges separately. This is in contrast
to traditional label propagation, where edges indicate similarity
exclusively \cite{zhu2005semi}.

We construct lexicons in 11 languages of varying morphological complexity.
We perform intrinsic evaluation of the quality of generated lexicons
obtained from either the universal dependency treebank or created manually by
humans (\S\ref{sec:intrinsic}).
We show that these automatically created lexicons provide useful
features in two  extrinsic NLP tasks which require identifying the contextually
plausible morphological and syntactic roles: morphological tagging
\cite{hajivc1998,hajivc2000} and syntactic dependency parsing
\cite{kubler2009}. We  obtain an average of 15.4\% and
5.3\% error reduction across 11 languages for morphological tagging and
dependency parsing respectively on a set of publicly available treebanks
(\S\ref{sec:extrinsic}). We anticipate that the lexicons thus created will be
useful in a variety of NLP problems.

\section{Graph Construction}
\label{sec:graph}

\begin{figure}[tb]
  \centering
  \includegraphics[width=1\columnwidth]{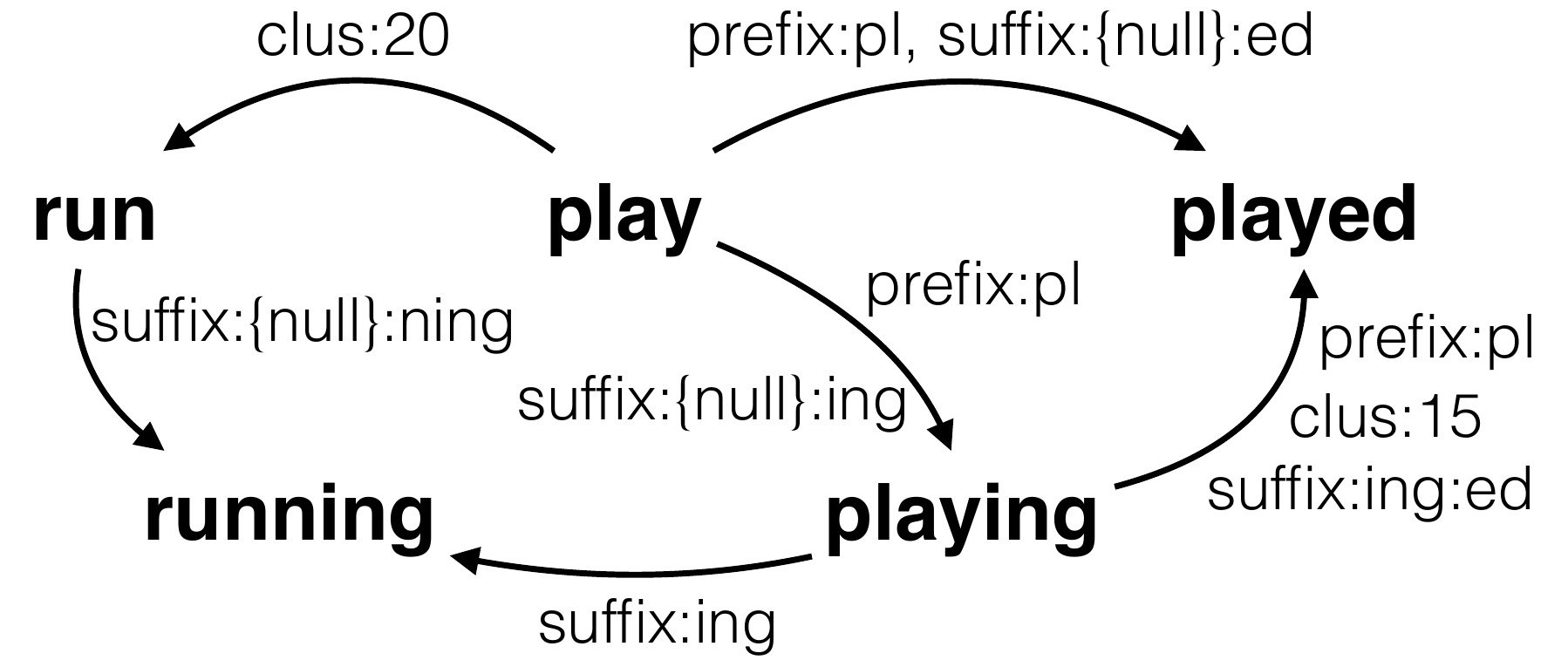}
  \caption{A subgraph from the complete graph of English showing
different kinds of features shared on the edges between words. Some possible
features/edges have been removed for enhancing clarity.}
  \label{fig:graph}
\end{figure}

The approach we take propagates information over lexical graphs
(\S\ref{sec:model}). In this section we describe how to construct the graph
that serves as the backbone of our model.
We construct a graph in
which nodes are word types and directed edges are present between nodes that
share \textit{one or more} features.
Edges between nodes denote that there might be a
relationship between the attributes of the two nodes, which we intend to learn.
As we want to keep our model language independent, we use
edge features that can be induced between words without using any language
specific tools. To this end, we describe three features in this section
that can be obtained using unlabeled corpora for any given
language.\footnote{Some of these features can cause the graph to become very
dense making label propagation prohibitive. We keep the size of the graph in
check by only allowing a word node to be connected to at most 100 other
(randomly selected) word nodes sharing one particular feature. This reduces
edges while still keeping the graph connected.}
Fig.~\ref{fig:graph} shows a subgraph of the full graph
constructed for English.

\paragraph{Word Clusters.} Previous work has shown that unlabeled text can be
used to induce unsupervised word clusters which can improve the performance of
many NLP tasks in different languages
\cite{clark:2003,koo:2008,turian:2010,faruqui:2010,tackstrom:2012,owoputi:2013}.
Word clusters capture semantic and syntactic similarities between words, for
example, \textit{play} and \textit{run} are present in the
same cluster. We obtain word clusters by using Exchange clustering algorithm
\cite{kneser1993improved,martin1998algorithms,uszkoreit2008distributed}
on large unlabeled corpus of every language.
As in \newcite{tackstrom:2012}, we use one year of news articles scrapped
from a variety of sources and cluster only the most frequent 1M words into
256 different clusters.
An edge was introduced for every word pair sharing the same word
cluster and a feature for the cluster is fired. Thus, there are 256 possible
cluster features on an edge, though in our case only a single one can fire.

\paragraph{Suffix \& Prefix.} Suffixes are often strong indicators of the
morpho-syntactic attributes of a word \cite{ratnaparkhi1996maximum,clark:2003}.
For example, in English, \textit{-ing} denotes gerund verb forms like,
\textit{studying}, \textit{playing} and \textit{-ed} denotes past tense like
\textit{studied}, \textit{played} etc. Prefixes like \textit{un-}, \textit{in-}
often denote adjectives. Thus we include both 2-gram and 3-gram suffix and
prefix as edge features.\footnote{We only include those suffix and prefix which
appear at least twice in the seed lexicon.} We introduce an edge between two
words sharing a particular suffix or prefix feature.

\paragraph{Morphological Transformations.} \newcite{soricut:2015} presented an
unsupervised method of inducing prefix- and suffix-based morphological
transformations between words using word embeddings. In their method,
statistically, most of the
transformations are induced between words with the same lemma (without using
any prior information about the word lemma). For example, their method
induces the transformation between \textit{played} and \textit{playing} as
\textit{suffix:ed:ing}. This feature indicates \textsc{Tense:Past} to turn off
and \textsc{Tense:Pres} to turn on.\footnote{Our model will
learn the following transformation: \textsc{Tense:Past}: 1 $\rightarrow$ -1, \textsc{Tense:Present}: -1 $\rightarrow$ 1 (\S\ref{sec:model}).}
\ignore{These are the only features that indicate a change in
the attributes (as these indicate the difference between two words) in
contrast to the other features which promote the propagation of same labels (as
they indicate similarity between words). This might seem counterintuitive
because generally label propagation connects nodes with similar attributes,
however, as we intend to learn the propagation function, such links can be higly
informative of attribute changes.}
We train the morphological transformation prediction tool of
\newcite{soricut:2015} on the news corpus (same as the one used for training
word clusters) for each language. An edge is introduced between two words
that exhibit a morphological transformation feature from one word to another
as indicated by the tool's output.

\paragraph{Motivation for the Model.}
To motivate our model, consider the words \textit{played} and \textit{playing}.
They have a common attribute \textsc{POS:Verb} but they differ in tense,
showing \textsc{Tense:Past} and \textsc{Tense:Pres} resp. Typical
graph-propagation algorithms model similarity \cite{zhu2005semi} and thus
propagate all attributes along the edges. However, we want to model if an
attribute should propagate or change across an edge. For example, having a
shared cluster feature, is an indication of similar POS tag
\cite{clark:2003,koo:2008,turian:2010}, but a
surface morphological transformation feature like \textit{suffix:ed:ing}
possibly indicates a change in the tense of the word. Thus, we will model
attributes propagation/transformation as a function of the features shared on
the edges between words.  The features described in this section
are specially suitable for languages that exhibit concatenative morphology, like
English, German, Greek etc. and might not work very well with languages that
exhibit non-concatenative morphology i.e, where root modification is highly
frequent like in Arabic and Hebrew.
However, it is important to note that our framework is not
limited to just the features described here, but can incorporate any arbitrary
information over word pairs (\S\ref{sec:future}).

\section{Graph-based Label Propagation}
\label{sec:model}

\begin{figure*}[tb]
  \centering
  \includegraphics[width=2\columnwidth]{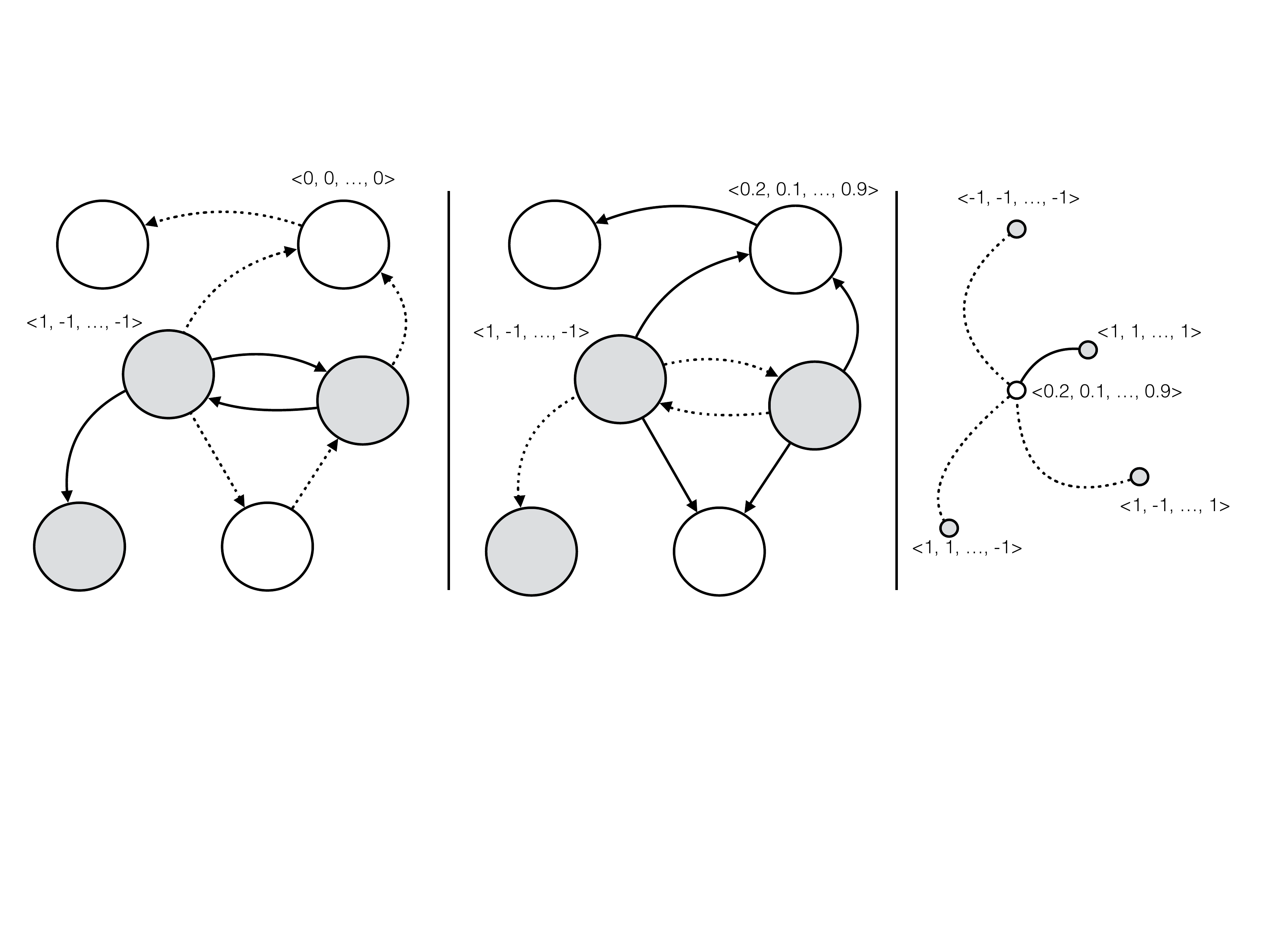}
  \caption{Word graph with edges between words showing the
    labeled (grey) and the unlabeled (white) word nodes. Only nodes connected
  via solid edges are visible to each other, dotted edges block visibility.
This figure demonstrates interaction between nodes during model estimation
(left), label propagation (center), and paradigm projection (right).
Attribute-value vectors of the words are shown in angled brackets. The solid
edge in the right figure shows the closest attribute paradigm to which the
empirical vector is projected.}
  \label{fig:schema}
\end{figure*}

We now describe our model.
Let $\mathcal{W} = \{w_1, w_2 , \ldots , w_{|\mathcal{W}|}\}$ be the vocabulary
with $|\mathcal{W}|$ words and
$\mathcal{A} = \{a_1, a_2 , \ldots , a_{|\mathcal{A}|}\}$ be the set of
lexical attributes that words in $\mathcal{W}$ can express; e.g.
$\mathcal{W}=\{played, playing, \ldots \}$ and
$\mathcal{A}=\{\textsc{Num:Sing}, \textsc{Num:Plur}, \textsc{Tense:Past},
\ldots\}$. Each word type $w\in \mathcal{W}$ is associated with a vector
$\vect{a}_w \in [-1,1]^{|\mathcal{A}|}$, where $a_{i,w}=1$ indicates that word
$w$ has attribute $i$ and $a_{i,w}=-1$ indicates its absence; values in between
are treated as degrees of uncertainty.  For example,
$\textsc{Tense:Past}_{played} = 1$ and $\textsc{Tense:Past}_{playing} =
-1$.\footnote{We constrain $a_{i,w} \in [-1, 1]$ as its easier to model
the flipping of an attribute value from $-1$ to $1$ as opposed to $[0, 1]$.}

The vocabulary $\mathcal{W}$ is divided
into two disjoint subsets, the labeled words $\mathcal{L}$ for which we know
their $\vect{a}_w$'s (obtained from seed lexicon)\footnote{We use labeled, seed,
and training lexicon to mean the same thing interchangeably.} and the unlabeled
words $\mathcal{U}$ whose attributes are unknown.
In general $|\mathcal{U}| \gg |\mathcal{L}|$.
The words in $\mathcal{W}$ are organized into a directed graph with edges
$\mathcal{E}$ between words.
Let, vector $\boldsymbol{\phi}(w, v) \in [0, 1]^{|\mathcal{F}|}$
denote the features on the directed edge between words $w$ and $v$, with $1$
indicating the presence and $0$ the absence of feature $f_k \in \mathcal{F}$,
where, $\mathcal{F} = \{f_1, f_2 , \ldots , f_{|\mathcal{F}|}\}$ are
the set of possible binary features shared between two words in the graph.
For example, the features on edges between \textit{played} and \textit{playing}
from Fig.~\ref{fig:graph} are:
\begin{equation*}
    \phi_{k}(played, playing)=
    \begin{cases}
      1, & \text{if} f_k = \text{suffix:ed:ing} \\
      1, & \text{if} f_k = \text{prefix:pl} \\
      0, & \text{if} f_k = \text{suffix:ly} \\
      \dots
    \end{cases}
  \end{equation*}

We seek to determine which subsets of $\mathcal{A}$ are valid for each word $w
\in \mathcal{W}$. We learn how a particular attribute of a node is a function of
\textit{that particular attribute} of its neighboring nodes and features on the
edge connecting them.
Let $a_{i, w}$ be an attribute of word $w$ and
let $\hat{a}_{i, w}$ be the empirical estimate of that attribute.
We posit that $\hat{a}_{i, w}$ can be estimated from the neighbors
$\mathcal{N}(w)$ of $w$ as follows:
\begin{equation}
\label{equ:emp}
\hat{a}_{i, w} = \mathrm{tanh}\left( \sum_{v \in \mathcal{N}(w)}
\left( \boldsymbol{\phi}(w, v) \cdot \boldsymbol{\theta_{i}} \right) \times
a_{i, v} \right)
\end{equation}
where, $\boldsymbol{\theta_{i}} \in \mathbb{R}^{|\mathcal{F}|}$
is weight vector of the edge features for estimating attribute $a_i$.
`$\cdot$' represents dot product betwen two vectors. We use $\textrm{tanh}$
as the non-linearity to make sure that $\hat{a}_{i, w} \in [-1, 1]$.
The set of such weights
$\boldsymbol{\theta} \in \mathbb{R}^{|\mathcal{A}|\times|\mathcal{F}|}$ for all
attributes are the model parameters that we learn.
Our graph resembles the Ising model, which is a lattice
model proposed for describing intermolecular forces \cite{ising1925beitrag}, and
eq.~\ref{equ:emp} solves the naive mean field approximation of the Ising model
\cite{wang2007semi}.

Intuitively, one can view the node to node message function from $v$ to
$w$: $\boldsymbol{\phi}(w, v) \cdot \boldsymbol{\theta_{i}} \times
a_{i, v} $ as either
(1) supporting the value $a_{i, v}$ when $\boldsymbol{\phi}(w, v) \cdot
\boldsymbol{\theta_{i}} > 0$; (2) inverting $a_{i, v}$
when $\boldsymbol{\phi}(w, v) \cdot \boldsymbol{\theta_{i}} < 0$; or
(3) dampening or neutering $a_{i, v}$ when $\boldsymbol{\phi}(w, v) \cdot
\boldsymbol{\theta_{i}} \approx 0$.
Returning to our motivation, if $w = played$ and $v = playing$, a feature
indicating the suffix substitution \textit{suffix:ed:ing} should
have a highly negative weight for \textsc{Tense:Past}, indicating a
change in value. This is because \textsc{Tense:Past} = -1 for \textit{playing},
and a negative value of $\boldsymbol{\phi}(w, v) \cdot \boldsymbol{\theta_{i}}$
will push it to positive for \textit{played}.

 It should be noted that this framework for constructing lexicons does not explicitly
distinguish between morpho-syntactic paradigms, but simply identifies all
possible attribute-values a word can take. If we consider an example like ``games'' and two
attributes, the syntactic part-of-speech, $\textsc{POS}$, and number, $\textsc{Num}$, games
can either be 1) $\{\textsc{POS:Verb}, \textsc{Num:Sing}\}$, as in \emph{John games the system}; or $\{\textsc{POS:Noun}, \textsc{Num:Plur}\}$, as in \emph{The games have started}.
Our framework will mereley return that all the above attribute-values are possible,
which implies that
the singluar noun and plural verb interpretations are valid. One possible way to
account for this is to make full morphological paradigms the ``attributes'' in or model.
But this leads to slower runtimes and sparser learning. We leave as future work extensions to
full paradigm prediction.

Our framework has three critical components, each described below:
(1) model estimation, i.e., learning $\boldsymbol{\theta}$;
(2) label propagation to $\mathcal{U}$; and optionally
(3) paradigm projection to known valid morphological paradigms.
The overall procedure is illustrated in Figure~\ref{fig:schema} and
made concrete in Algorithm~\ref{algo:prop}.

\begin{algorithm}[tb]
 \DontPrintSemicolon
\LinesNumbered
 \KwData{$\mathcal{W}$, $\mathcal{L}$, $\mathcal{U}$, $\mathcal{A}$,
 $\mathcal{F}$, $\mathcal{P}$}
 \KwResult{$\boldsymbol{\theta}$, labeled $\mathcal{U}$}
 \tcp{model estimation}
 \While{not convergence} {
   \For{$w \in \mathcal{L}$}{
    loss $\leftarrow \Vert \vect{a}_w - \hat{\vect{a}}_w \Vert_2^2$\;
    Update $\boldsymbol{\theta}$ using $\frac{\partial \mathrm{loss}}{\partial \boldsymbol{\theta}}$\;
   }
 }
 \tcp{label propagation}
 \While{not convergence} {
   \For{$w \in \mathcal{U}$}{
     $\vect{a}_w \leftarrow \hat{\vect{a}}_w$\;
   }
 }
 \tcp{paradigm projection}
 \For{$w \in \mathcal{U}$}{
   mindist $\leftarrow \infty$, closest $\leftarrow \emptyset$\;
   \For{$\vect{p} \in \mathcal{P}$}{
     dist $\leftarrow \Vert \vect{a}_w - \vect{p} \Vert_2^2$\;
     \If{dist $<$ mindist} {
       mindist $\leftarrow$ dist, closest $\leftarrow \vect{p}$\;
     }
   }
   $\vect{a}_w  \leftarrow$ closest\;
 }
 \caption{Graph-based semi-supervised label propagation algorithm.}
 \label{algo:prop}
\end{algorithm}

\subsection{Model Estimation}
\label{sec:train}

We estimate all individual elements of an attribute vector
using eq.~\ref{equ:emp}. We define loss as the squared loss between the
empirical and observed attribute vectors on every labeled node in the graph,
thus the total loss can be computed as:
\begin{equation}
\label{equ:loss}
\sum_{w \in \mathcal{L}} \Vert \mathbf{a}_{w} - \hat{\mathbf{a}}_{w} \Vert_2^2
\end{equation}

We train the edge feature weights $\boldsymbol{\theta}$ by minimizing the loss
function in eq.~\ref{equ:loss}. In this step, we only use labeled nodes and the
edge connections between labeled nodes.
As such, this is strictly a supervised learning setup.
We minimize the loss function using online adaptive gradient descent
\cite{duchi2011adaptive} with $\ell_2$ regularization
on the feature weights $\boldsymbol{\theta}$. This is the first step in
Algorithm~\ref{algo:prop} (lines 1--4).

\subsection{Label Propagation}
\label{sec:prop}

In the second step, we use the learned weights of the edge features to estimate
the attribute values over unlabeled nodes iteratively.
The attribute vector of all unlabeled words is initialized to null,
$\forall w \in \mathcal{U}, \vect{a}_w = \langle 0, 0, \ldots, 0 \rangle$.
In every iteration, an
unlabeled node estimates its empirical attributes by looking at the
corresponding attributes of its labeled and unlabeled neighbors
using eq.~\ref{equ:emp}, thus this is the
semi-supervised step. We stop after the squared euclidean distance between the
attribute vectors at two consecutive iterations for a node becomes less than 0.1
(averaged over all unlabeled nodes). This is the second step in
Algorithm~\ref{algo:prop} (lines 5--7). After convergence, we can
directly obtain attributes for a word by thresholding:
a word $w$ is said to possess an attribute $a_i$ if $a_{i, w} > 0$.

\subsection{Paradigm Projection}
\label{sec:proj}

Since a word can be labeled with multiple lexical attributes, this is
a multi-label classification problem. For such a task, several advanced methods
that take into account the correlation between attributes have been proposed
\cite{ghamrawi:2005,tsoumakas:2006,furnkranz:2008,read:2011},
here we have adopted the \textit{binary relevance method} which trains a
classifier for every attribute independently of the other attributes, for its
simplicity \cite{godbole:2004,zhang:2005}.

However, as the decision for the presence of an attribute over a word is
independent of all the other attributes, the final set of attributes
obtained for a word in \S\ref{sec:prop} might not be a valid
paradigm.\footnote{A paradigm is defined as a set of attributes.}
For example, a word cannot only exhibit the two
attributes \textsc{POS:Noun} and \textsc{Tense:Past}, since the presence of
the tense attribute implies \textsc{POS:Verb} should also be true.
Further, we want to utilize the
inherent correlations between attribute labels to obtain better solutions.
We thus present an alternative, simpler method to account for this problem.
To ensure that we obtain a valid attribute paradigm, we project the empirical
attribute vector obtained after propagation to the space of all valid
paradigms.

We first collect all observed  and thus valid attribute paradigms from
the seed
lexicon ($\mathcal{P} = \{\vect{a}_w | w \in \mathcal{L} \}$).
We replace the empirical attribute vector obtained in \S\ref{sec:prop} by
a valid attribute paradigm vector which is nearest to it according to
euclidean distance. This projection step is inspired from the decoding step in
label-space transformation approaches to multilabel classification
\cite{hsu2009multi,ferng2011multi,zhang2011multi}. This is the last step in
Algorithm~\ref{algo:prop} (lines 8--14). We investigate for each language
if paradigm projection is helpful (\S\ref{sec:deplex}).

\section{Intrinsic Evaluation}
\label{sec:intrinsic}

\begin{table}[!tb]
  \centering
  \begin{tabular}{|l|rrr|rr|r|}
  \hline
   & $|\mathcal{L}|$ & $|\mathcal{W}|$ & $|\mathcal{E}|$ & $|\mathcal{A}|$ & $|\mathcal{P}|$ & Prop \\
   & (k) & (k) & (m) & & & (k)\\
  \hline
  eu & 3.4 & 130 & 13 & 83 & 811& 118 \\
  bg & 8.1 & 309 & 27 & 57 & 53& 285\\
  hr & 6.9 & 253 & 26 & 51 & 862& 251 \\
  cs & 58.5 & 507 & 51 & 122 & 4,789& 403 \\
  da & 5.9 & 259 & 26 & 61 & 352 & 246 \\
  en & 4.1 & 1,006 & 100 & 50 & 412 & 976\\
  fi & 14.4 & 372 & 30 & 100 & 2,025 & 251\\
  el & 3.9 & 358 & 26 & 40 & 409 & 236\\
  hu & 1.9 & 451 & 25 & 85 & 490 & 245\\
  it & 8.5 & 510 & 28 & 52 & 568 & 239\\
  sv & 4.8 & 305 & 26 & 41 & 265 & 268\\
  \hline
  \end{tabular}
  \caption{Graph statistics for different languages, showing the approximate
number of labeled seed nodes ($|\mathcal{L}|$), labeled and unlabeled nodes
($|\mathcal{W}|$), edges between words ($|\mathcal{E}|$), the number
of unique attributes ($|\mathcal{A}|$), attribute paradigms ($|\mathcal{P}|$)
and size of the constructed lexicon (Prop). k: thousands, m: millions.}
  \label{tab:lexicons}
\end{table}

To ascertain how our graph-propagation framework predicts morphological
attributes for words, we provide an intrinsic
evaluation where we compare predicted attributes to gold lexicons that have
been either read off from a treebank or derived manually.

\subsection{Dependency Treebank Lexicons}
\label{sec:deplex}

The universal dependency treebank
\cite{mcdonald:2013,de2014universal,agic:2015} contains dependency
annotations for sentences and morpho-syntactic annotations for words in context
for a number of languages.\footnote{We use version 1.1
released in May 2015.}
A word can display different attributes depending
on its role in a sentence.
In order to create morpho-syntactic lexicon for every language, we take the
union of all the attributes that the word realizes in the entire
treebank.
Although, it is possible that this lexicon might not contain all realizable
attributes if a particular attribute or paradigm is not seen in the
treebank (we address this issue in \S\ref{sec:manual}).
The utility of evaluating against treebank derived lexicons is that it allows us
to evaluate on a large set of languages. In particular, in the universal
dependency treebanks v1.1 \cite{agic:2015}, 11 diverse languages contain the
morphology layer, including Romance, Germanic and Slavic languages plus
isolates like Basque and Greek.

We use the train/dev/test set of the treebank to create
training (seed),\footnote{We only include those words in the seed lexicon that
occur at least twice in the training set of the treebank.} development and test
lexicons for each language. We exclude
words from the dev and test lexicon that have been seen in seed lexicon.
For every language, we create a graph with the features described in
\S\ref{sec:graph} with words in the seed lexicon as labeled nodes.
The words from development and test set are included as unlabeled nodes
for the propagation stage.\footnote{Words from the news corpus used for word
clustering are also used as unlabeled nodes.}
Table~\ref{tab:lexicons} shows statistics about the constructed graph for
different languages.\footnote{Note that the size of the constructed lexicon
(cf. Table~\ref{tab:lexicons}) is always less than or equal to the total number
of unlabeled nodes in the graph because some unlabeled nodes are not able to
collect enough mass for acquiring an attribute i.e,
$\forall a \in \mathcal{A} : a_w < 0$ and thus they remain unlabeled (cf.
\S\ref{sec:prop}).}

We perform feature selection and hyperparameter tuning by optimizing prediction
on words in the development lexicon and then report results on the test lexicon.
The decision whether paradigm projection (\S\ref{sec:proj}) is useful or not
is also taken by tuning performance on the development lexicon.
Table~\ref{tab:feats} shows the features that were selected for each language.
Now, for every word in the test lexicon we obtain predicted lexical attributes
from the graph. For a given attribute, we count the number of words for which
it was correctly predicted (true positive), wrongly predicted (false positive)
and not predicted (false negative). Aggregating these counts over all attributes
($\mathcal{A}$),
we compute the micro-averaged $F_1$ score and achieve 74.3\% on an average
across 11 languages (cf. Table~\ref{tab:intrinsic}). Note that this
systematically underestimates performance due to the effect of missing
attributes/paradigms that were not observed in the treebank.

\paragraph{Propagated Lexicons.} The last column in
Table~\ref{tab:lexicons} shows the number of words in
the propagated lexicon, and the first column shows the number of words in
the seed lexicon. The ratio of the size of propagated and seed lexicon is
different across languages, which presumably depends on how densely connected
each language's graph is. For example, for English the propagated lexicon is
around $240$ times larger than the seed lexicon, whereas for Czech, its $8$
times larger. We can individually tune how densely connected graph we want
for each language depending on the seed size and feature sparsity,
which we leave for future work.

\paragraph{Selected Edge Features.}
The features most frequently selected across all the languages are the
word cluster and the surface morphological transformation features.
This essentially
translates to having a graph that consists of small connected components
of words having the same lemma (discovered in an unsupervised manner) with
semantic links connecting such components using word cluster features. Suffix
features are useful for highly inflected languages like Czech and Greek,
while the prefix feature is only useful for Czech.
Overall, the selected edge features for different languages
correspond well to the morphological structure of these languages
\cite{wals-26}.

\begin{table}[!tb]
  \centering
  \begin{tabular}{|c|cccc|c|}
  \hline
  & Clus & Suffix & Prefix & MorphTrans & Proj\\
  \hline
  eu & \checkmark & & & \checkmark & \checkmark \\
  bg & \checkmark & & & \checkmark & \\
  hr & \checkmark & \checkmark & & & \checkmark \\
  cs & \checkmark & \checkmark & \checkmark & \checkmark & \checkmark \\
  da & \checkmark & & & \checkmark & \checkmark \\
  en & \checkmark & & & \checkmark & \checkmark \\
  fi & \checkmark & & & \checkmark &  \\
  el & \checkmark & \checkmark & & \checkmark & \checkmark \\
  hu & \checkmark & & & \checkmark & \checkmark \\
  it & \checkmark & & & \checkmark & \checkmark \\
  sv & \checkmark & \checkmark & & & \checkmark \\
  \hline
  \end{tabular}
  \caption{Features selected and the decision of paradigm
  projection (Proj) tuned on the development lexicon for each language.
 \checkmark denotes a selected feature.}
  \label{tab:feats}
\end{table}

\paragraph{Corpus Baseline.}
We compare our results to a corpus-based method of obtaining morpho-syntactic
lexicons. We hypothesize that if we use a morphological tagger of reasonable
quality to tag the entire wikipedia corpus of a language and take the union of
all the attributes for a word
type across all its occurrences in the corpus, then we can acquire all possible
attributes for a given word. Hence, producing a lexicon of reasonable quality.
\newcite{moore:2015:EMNLP} used this technique to obtain a high
quality tag dictionary for POS-tagging.
We thus train a morphological tagger (detail in
\S\ref{sec:morph}) on the training portion of the
dependency treebank and use it to tag the entire wikipedia corpus. For every
word, we add an attribute to the lexicon if it has been seen at least $k$
times for the word in the corpus, where $k \in [2, 20]$.
This threshold on the frequency of the word-attribute pair helps prevent
noisy judgements. We tune $k$ for each
language on the development set and report results on the test set in
Table~\ref{tab:intrinsic}. We call this method the Corpus baseline. It can be
seen that for every language we outperform this baseline, which on average has
an $F_1$ score of 67.1\%.

\begin{table}[!tb]
  \centering
  \begin{tabular}{|c|r||r|r|}
  \hline
  & words & Corpus & Propagation\\
  \hline
  eu & 3409 & 54.0 & 57.5 \\
  bg & 2453 & 66.4 & 73.6 \\
  hr & 1054 & 69.7 & 71.6 \\
  cs & 14532 & 79.1 & 80.8 \\
  da & 1059 & 68.2 & 78.1 \\
  en & 3142 & 57.2 & 72.0 \\
  fi & 2481 & 58.2 & 68.2 \\
  el & 1093 & 72.2 & 82.4 \\
  hu & 1004 & 64.9 & 70.9 \\
  it & 1244 & 78.8 & 81.7 \\
  sv & 3134 & 69.8 & 80.7 \\
  \hline
  avg. & 3146 & 67.1 & 74.3\\
  \hline
  \end{tabular}
  \caption{Micro-averaged $F_1$ score (\%) for prediction of lexical attributes on
  the test set using our propagation algorithm (Propagation) and the
  corpus-based baseline (Corpus). Also, shown are the no. of words
  in test set.}
  \label{tab:intrinsic}
\end{table}

\subsection{Manually Curated Lexicons}
\label{sec:manual}

\begin{table}[!tb]
  \centering
  \begin{tabular}{|l|r|r|}
  \hline
  & words & $F_1$\\
  \hline
  cs & 115,218 & 87.5 \\
  fi & 39,856 & 71.9 \\
  hu & 135,386 & 79.7 \\
  \hline
  avg. & 96,820 & 79.7\\
  \hline
  \end{tabular}
  \caption{Micro-averaged $F_1$ score (\%) for prediction of
  lexical attributes on the test lexicon of human-curated lexicons.}
  \label{tab:reallex}
\end{table}

We have now showed that its possible to automatically
construct large lexicons from smaller seed lexicons. However, the seed lexicons
used in \S\ref{sec:deplex} have been artifically constructed from aggregating
attributes of word types over the treebank. Thus, it can be argued that these
constructed lexicons might not be complete i.e, the lexicon might not exhibit
all possible attributes for a given word. On the other hand, manually curated
lexicons are unavailable for many languages, inhibiting proper evaluation.

To test the utility of our approach on manually curated lexicons, we investigate
publicly available lexicons for Finnish \cite{pirinen2011}, Czech
\cite{hajivc1998} and Hungarian \cite{tron2006}. We eliminate numbers and
punctuation from all lexicons. For each of these languages, we select 10000
words for training and the rest of the word types for evaluation.
We train models obtained in
\S\ref{sec:deplex} for a given language using suffix, brown and morphological transformation
features with paradigm projection. The only difference is the source
of the seed lexicon and test set. Results are reported in
Table~\ref{tab:reallex} averaged over 10 different randomly selected seed set
for every language. For each language we obtain more than $70$\% $F_1$ score
and on an average obtain $79.7$\%. Critically, the $F_1$ score on human
curated lexicons is higher for each language than the treebank constructed
lexicons, in some cases as high as $9$\% absolute. This shows that the average
$74.3$\% $F_1$ score across all 11 languages is likely underestimated.

\section{Extrinsic Evaluation}
\label{sec:extrinsic}

We now show that the automatically generated lexicons provide informative
features that are useful in two downstream NLP tasks: morphological tagging
(\S\ref{sec:morph}) and syntactic dependency parsing (\S\ref{sec:parse}).

\subsection{Morphological Tagging}
\label{sec:morph}

\begin{table}[!tb]
  \centering
  \begin{tabular}{|c|c|}
  \hline
  word & exchange cluster$^*$ \\
  lowercase(word) &  capitalization \\
  \{1,2,3\}-g suffix$^*$ & digit \\
  \{1,2,3\}-g prefix$^*$ & punctuation \\
  \hline
  \end{tabular}
  \caption{Features used to train the morphological tagger on the universal dependency treebank. $^*$:on for word offsets \{-2, -1, 0, 1, 2\}. Conjunctions of the above
  are also included.}
  \label{tab:morphfeat}
\end{table}

\begin{table}[!tb]
  \centering
  \begin{tabular}{|l|rr|r|}
  \hline
  & None & Seed & Propagation \\
  \hline
  eu & 84.1 & 84.4 & \textbf{85.2} \\
  bg & 94.2 & 94.6 & \textbf{95.9} \\
  hr & 92.5 & \textbf{93.6} & 93.2 \\
  cs & 96.8 & 97.1 & 97.1 \\
  da & 96.4 & 97.1 & \textbf{97.3} \\
  en & 94.4 & 94.7 & \textbf{94.8} \\
  fi & 92.8 & 93.6 & \textbf{94.0} \\
  el & 93.4 & \textbf{94.6} & 94.2 \\
  hu & 91.7 & 92.3 & \textbf{93.5}\\
  it & 96.8 & 97.1 & 97.1 \\
  sv & 95.4 & 96.5 & 96.5 \\
  \hline
  avg. & 93.5 & 94.2 & \textbf{94.5} \\
  \hline
  \end{tabular}
  \caption{Macro-averaged $F_1$ score (\%) for morphological tagging: without
  using any lexicon (None), with seed lexicon (Seed), with propagated lexicon
(Propagation).}
  \label{tab:morph}
\end{table}

Morphological tagging is the task of assigning a morphological reading to a token in context.
The morphological reading consists of features such as part of speech, case, gender, person,
tense etc. \cite{oflazer1994tagging,hajivc1998}. The model we use is a standard
atomic sequence classifier, that classifies the morphological bundle for each word
independent of the others (with the exception of features derived from these words).
Specifically, we use a linear SVM model classifier with hand tuned features.
This is similar to commonly used analyzers like SVMTagger \cite{marquez2004}
and MateTagger \cite{bohnet2012}.

Our taggers are trained in a language independent manner
\cite{hajivc2000,smith2005context,muller2013efficient}. The list of features
used in training the tagger are listed in Table~\ref{tab:morphfeat}.
In addition to the standard features, we use the
morpho-syntactic attributes present in the lexicon for every word as features in the tagger.
As shown in \newcite{muller:2015}, this is typically the most important feature
for morphological tagging, even more useful than clusters or word embeddings.
While predicting the contextual morphological tags for a given word, the
morphological attributes present in the lexicon for the current word, the previous word and the
next word are used as features.

We use the same 11 languages from the universal dependency treebanks
\cite{agic:2015} that contain morphological tags to train and evaluate the
morphological taggers. We use the pre-specified train/dev/test splits that come
with the data.
Table~\ref{tab:morph} shows the macro-averaged $F_1$ score over all attributes
for each language on the test lexicon. The three columns show the $F_1$ score of
the tagger when no lexicon is used; when the seed lexicon derived from the
training data is used; and when label propagation is applied.

Overall, using
lexicons provides a significant improvement in accuracy, even when just using
the seed lexicon. For 9 out of 11 languages, the highest accuracy is
obtained using the lexicon derived from graph propagation.
In some cases the gain is quite substantial, e.g., 94.6\% $\rightarrow$ 95.9\%
for Bulgarian. Overall there is 1.0\% and 0.3\% absolute improvement over the
baseline and seed resp., which corresponds roughly to a 15\% and 5\%
relative reduction in error. It is not surprising that the seed lexicon
performs on par with the derived lexicon for some languages, as it is
derived from the training corpus, which likely contains the most frequent words
of the language.

\subsection{Dependency Parsing}
\label{sec:parse}

\begin{table}[!tb]
  \centering
  \begin{tabular}{|l|rr|r|}
  \hline
  & None & Seed & Propagation \\
  \hline
  eu & 60.5 & 62.3 & \textbf{62.9} \\
  bg & 78.3 & 78.8 & \textbf{79.3} \\
  hr & 72.8 & 74.7 & 74.7\\
  cs & 78.3 & 78.4 & 78.4 \\
  da & 67.5 & 69.4 & \textbf{70.1} \\
  en & 74.4 & 74.1 & \textbf{74.4} \\
  fi & 66.1 & 67.4 & \textbf{67.9}\\
  el & 75.0 & 75.6 & \textbf{75.8}\\
  hu & 67.6 & 69.0 & \textbf{71.1}\\
  it & 82.4 & 82.8 & \textbf{83.1}\\
  sv & 69.7 & 70.1 & 70.1\\
  \hline
  avg. & 72.0 & 73.0 & \textbf{73.5}\\
  \hline
  \end{tabular}
  \caption{Labeled accuracy score (LAS, \%) for dependency parsing:
  without using any lexicon (None), with seed (Seed),
  with propagated lexicon (Propagation).}
  \label{tab:parse}
\end{table}

We train dependency parsers for the same 11 universal dependency
treebanks that contain the morphological layer \cite{agic:2015}.
We again use the supplied train/dev/test split of the
dependency treebank to develop the models. Our parsing model is the
transition-based parsing system of \newcite{zhang2011} with
identical features and a beam of size 8.

We augment the features of \newcite{zhang2011} in two ways:
using the context-independent morphological attributes present in the different
lexicons; and using the corresponding morphological taggers from
\S\ref{sec:morph} to generate context-dependent attributes.
For each of the above two kinds of features, we fire the attributes
for the word on top of the stack and the two words on at the front of
the buffer. Additionally we take the cross product of these
features between the word on the top of the stack and at the front of
the buffer.

Table~\ref{tab:parse} shows
the labeled accuracy score (LAS) for all languages.
Overall, the generated lexicon gives an improvement of absolute $1.5\%$ point
over the baseline (5.3\% relative reduction in error) and
$0.5\%$ over the seed lexicon on an average across 11 languages.
Critically this improvement holds for 10/11 languages
over the baseline and 8/11 languages over the system that
uses seed lexicon only.

\section{Further Analysis}

\begin{table*}[!tb]
  \centering
  \begin{tabular}{|l|l|l|}
  \hline
  & Word & Attributes \\
  \hline
  \multirow{3}{*}{en}
  & study (seed) & POS:Verb, VForm:Fin, Mood:Ind, Tense:Pres, Num:Sing,
  POS:Noun \\
  & studied & POS:Verb, VForm:Fin, Mood:Ind, Tense:Past, VForm:Part\\
  & taught & POS:Verb, VForm:Fin, Mood:Ind, Tense:Past, VForm:Part, Voice:Pass\\
  \hline
  \multirow{3}{*}{it} & tavola (seed) & POS:Noun, Gender:Fem, Num:Sing \\
  & tavoli & POS:Noun, Gender:Masc, Num:Plur \\
  & divano & POS:Noun, Gender:Masc, Num:Sing \\
  \hline
  \end{tabular}
  \caption{Attributes induced for words which are semantically or
  syntactically related to a word in the seed lexicon for English and Italian.
 }
  \label{tab:examples}
\end{table*}

In this section we further investigate our model and results in detail.

\begin{figure}[tb]
  \centering
  \includegraphics[width=0.9\columnwidth]{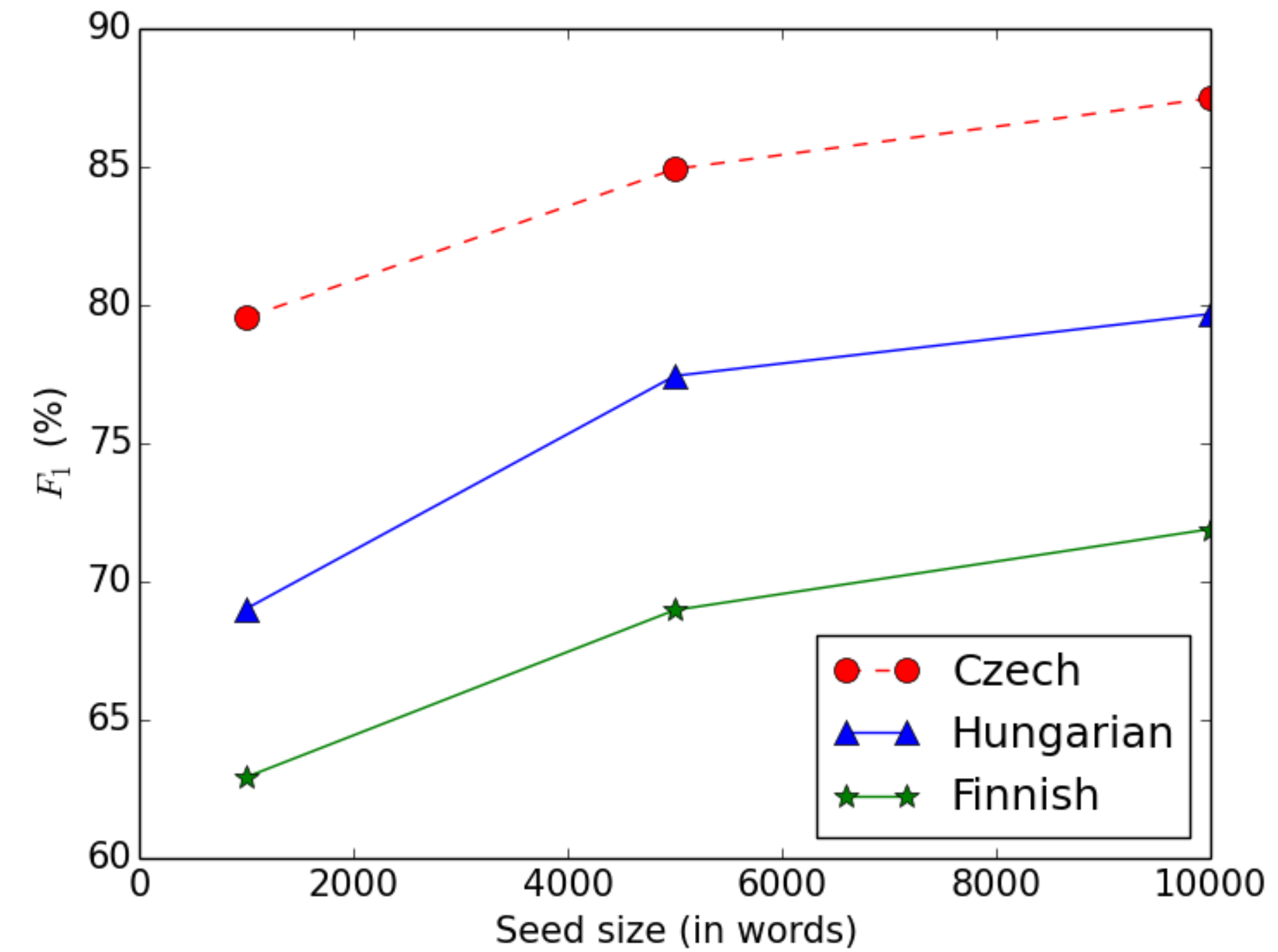}
  \caption{Micro-average $F_1$ score on test lexicon while using varying
seed sizes for cs, hu and fi.}
  \label{fig:plot}
\end{figure}

\paragraph{Size of seed lexicon.}
We first test how the size of the seed lexicon affects performance of
attribute prediction on the test set. We use the manually constructed lexicons
described in \S\ref{sec:manual} for experiments.
For each language, instead of using the
full seed lexicon of 10000 words, we construct subsets of this lexicon by
taking 1000 and 5000 randomly sampled words. We then train models
obtained in \S\ref{sec:deplex} on these lexicons and plot the performance
on the test set in Figure~\ref{fig:plot}. On average across three languages,
we observe that the absolute performance improvement from 1000
to 5000 seed words is $\approx$10\% whereas it reduces to $\approx$2\%
from 5000 to 10000 words.

\begin{table}[!tb]
  \centering
  \begin{tabular}{|lr|lr|}
  \hline
  \textsc{VForm:Ger}& & \textsc{Num:Plur} &\\
  \hline
  Clus:105 & + & Clus:19 & + \\
  Clus:77 & + & Clus:97 & + \\
  Clus:44 & + & Clus:177 & + \\
  \hline\hline
  suffix:ing:\{null\} & - & suffix:ies:y & - \\
  suffix:ping:\{null\} & - & suffix:gs:g & - \\
  suffix:ing:er & - & suffix:ons:on &  - \\
  \hline
  \end{tabular}
  \caption{Highest (upper half) and lowest (lower half) weighted features
  (with their sign) for predicting a given attribute of English words.}
  \label{tab:featana}
\end{table}

\paragraph{Feature analysis.} Table~\ref{tab:featana} shows the highest and the
lowest weighted features for predicting a given attribute of English words.
The highest weighted features for both \textsc{VForm:Ger} and
\textsc{Num:Plur} are word clusters, indicating that word clusters exhibit
strong syntactic and semantic coherence. More interestingly, it can be seen
that for predicting \textsc{VForm:Ger} i.e, continuous verb forms,
the lowest weighted features are those morphological transformations that
substitute ``ing'' with something else. Thus, if there exists an edge between
the words \textit{studying} and \textit{study}, containing the feature:
suffix:ing:\{null\}, the model would correctly predict that \textit{studying}
is \textsc{VForm:Ger} as \textit{study} is not so and the negative feature
weight can flip the label values. The same observation holds true for
\textsc{Num:Plur}.

\begin{table}[!tb]
  \centering
  \begin{tabular}{|l|ccc|}
  \hline
  & cs & hu & fi \\
  \hline
  S + C + MT & \textbf{87.5} & \textbf{79.9} & \textbf{71.6} \\
  S + C & 86.5 & 78.8 & 68.2 \\
  S + MT & 85.7 & 77.0 & 68.7 \\
  C + MT & 75.7 & 57.4 & 62.2 \\
  S + C + MT + P & 86.7 & 66.0 & 61.3\\
  \hline
  \end{tabular}
  \caption{Feature ablation study for induced lexicons evaluated on manually
curated gold lexicons. Reported scores are micro-averaged $F_1$ score (\%) for
prediction of lexical attributes. S = suffix; P = prefix; C = clusters; and MT
= morphological transformations.}
  \label{tab:ablation}
\end{table}

\paragraph{Feature ablation.}
One key question is which of the features in our graph are important for projecting
morphological attribute-values. Table~\ref{tab:feats} suggests that this is language
specific, which is intuitive, as morphology can be
represented more or less regularly through the surface form depending on the language. To understand this, we did a feature ablation study for the three languages with manually curated
lexicons (\S\ref{sec:manual}) using the same feature set as before: clusters, suffix and morphological transformations with paradigm projection. We then leave out each feature to measure how
performance drops. Unlike \S\ref{sec:manual}, we do not average over 10 runs but use a
single static graph where features (edges) are added or removed as necessary.

Table~\ref{tab:ablation} contains the results. Critically, all features are required for
top accuracy across all languages and leaving out suffix features has the most
detrimental effect. This is not surprising considering all three language primarily express
morphological properties via suffixes. Furthermore, suffix features help to
connect the graph and assist label propagation. Note that the importance of
suffix features here is in contrast to the evaluation on treebank derived
lexicons in \S\ref{sec:deplex}, where suffix features were only selected
for 4 out of 11 languages based on the development data
(Table~\ref{tab:feats}), and not for Hungarian and Finnish.
This could be due to the nature of the lexicons derived from treebanks versus complete lexicons
constructed by humans.

Additionally, we also added back prefix features and found that for all
languages, this resulted in a drop in accuracy, particularly for Finnish and
Hungarian. The primary reason for this is that prefix features often create
spurious edges in the graph. This in and of itself is not a problem for our
model, as the edge weights should learn to discount this feature. However, the
fact that we sample edges to make inference tractable means that more
informative edges could be dropped in favor of those that are only connected
via a prefix features.

\paragraph{Prediction examples.} Table~\ref{tab:examples} shows examples of
predictions made by our model for English and Italian. For each language, we
first select a random word from the seed lexicon, then we pick one syntactic
and one semantically related word to the selected word from the set of unlabeled
words. For e.g., in Italian \textit{tavola} means \textit{table}, whereas
\textit{tavoli} is the plural form and \textit{divano} means \textit{sofa}. We
correctly identify attributes for these words.

\section{Related Work}

We now review the areas of related work.

\paragraph{Lexicon generation.}
\newcite{eskander:2013} construct morpho-syntactic lexicons by incrementally
merging inflectional classes with shared morphological features.
Natural language lexicons have often been created from smaller seed lexcions
using various methods. \newcite{Thelen:2002} use patterns extracted over a
large corpus to learn semantic lexicons from smaller seed lexicons using
bootstrapping.
\newcite{alfonseca:2010} use distributional similarity scores across instances
to propagate attributes using random walks over a graph.
\newcite{das-smith12} learn potential
semantic frames for unknown predicates by expanding a seed frame lexicon.
Sentiment lexicons containing semantic polarity labels for
words and phrases have been created using bootstrapping and graph-based learning
\cite{banea2008,Mohammad:2009,velikovich2010,takamura,Lu:2011}.

\paragraph{Graph-based learning.}
In general, graph-based semi-supervised learning is heavily used in NLP
\cite{talukdar:2013,subramanya2014graph}.
Graph-based learning has been used for class-instance acquisition
\cite{talukdar2010experiments}, text classification \cite{subramanya2008soft},
summarization \cite{erkan2004lexrank}, structured prediction problems
\cite{Subramanya:2010,das-petrov:2011:ACL-HLT2011,garrette2013real} etc.
Our work differs from most of these approaches in that we specifically
learn how different features shared between the nodes can correspond to either
the propagation of an attribute or an inversion of the attribute value
(cf. equ~\ref{equ:emp}). In terms of the capability of inverting an attribute
value, our method is close to \newcite{goldberg2007dissimilarity}, who present
a framework to include dissimilarity between nodes and
\newcite{talukdar2012acquiring}, who learn which edges can be excluded for
label propagation. In terms of featurizing the edges, our work resembles
previous work which measured similarity between nodes in terms of similarity
between the feature types that they share \cite{muthu:2011,saluja:2013}.
Our work is also related to graph-based metric learning, where
the objective is to learn a suitable distance
metric between the nodes of a graph for solving a given problem
\cite{weinberger2005,dhillon:2012}.

\paragraph{Morphology.}
High morphological complexity exacerbates the
problem of feature sparsity in many NLP applications and leads to poor
estimation of model parameters, emphasizing the need of morphological analysis.
Morphological analysis encompasses fields like morphological segmentation
\cite{creutz2005unsupervised,demberg2007language,snyder2008unsupervised,poon2009unsupervised,narasimhan2015unsupervised}, and inflection
generation \cite{yarowsky:00,wicentowski:04}. Such models of segmentation
and inflection generation are used to better understand the meaning and
relations between words. Our task is complementary to the task of
morphological paradigm generation. Paradigm generation requires
generating all possible morphological forms of a given base-form according to
different linguistic transformations
\cite{de:11,ddn:13,ahlberg:14,ahlberg:15,kondrak:15:inflection,faruqui:2016:infl}, whereas our
task requires identifying linguistic transformations between two different word
forms.

\paragraph{Low-resourced languages.}  Our algorithm can be used to
generate morpho-syntactic lexicons for low-resourced languages, where the seed
lexicon can be constructed, for example, using crowdsourcing
\cite{Callison-Burch:2010,Irvine:201}. Morpho-syntactic resources have been
developed for east-european languages like Slovene
\cite{dzeroski:00,erjavec2004multext},
Bulgarian \cite{simov2004language} and highly agglutinative languages like
Turkish \cite{sak2008turkish}. Morpho-syntactic lexicons are crucial components
in acousting modeling and automatic speech recognition, where they have been
developed for low-resourced languages \cite{huet2008,besacier2014automatic}.

One alternative method to extract morphosyntactic lexicons is via parallel data \cite{das-petrov:2011:ACL-HLT2011}. However, such methods assume that both the source
and target langauges are isomorphic with respect to morphology. This can be the case with
attributes like coarse part-of-speech or case, but is rarely true for other attributes like
gender, which is very language specific.

\section{Future Work}
\label{sec:future}

There are three major ways in which the current model can be possibly improved.

\paragraph{Joint learning and propagation.} In the
current model, we are first learning the weights in a supervised manner
(\S\ref{sec:train}) and then propagating labels across nodes in a
semi-supervised step with fixed feature weights (\S\ref{sec:prop}). These
can also be performed jointly: perform one iteration of weight learning,
propagate labels using these weights, perform another iteration of weight
learning assuming empirical labels as gold labels and continue to learn and
propagate until convergence. This joint learning would be slower than
the current approach as propagating labels across the graph is an expensive
step.

\paragraph{Multi-label classifcation.} We are currently using the
\textit{binary relevance method} which trains a binary classifier for every
attribute independently \cite{godbole:2004,zhang:2005} with paradigm
projection as a post-processing step (\S\ref{sec:proj}). Thus we are accounting
for attribute correlations only at the end. We can instead model such
correlations as constraints during the learning step to obtain better solutions
\cite{ghamrawi:2005,tsoumakas:2006,furnkranz:2008,read:2011}.

\paragraph{Richer feature set.}
In addition our model can benefit from a richer set of features. Word
embeddings can be used to connect word node which are similar in meaning
\cite{mikolov-yih-zweig:2013:NAACL}. We can use existing morphological
segmentation tools to discover the morpheme and inflections of a word
to connect it to word with similar inflections which might be better than the
crude suffix or prefix features.
We can also use rich lexical resources
like Wiktionary\footnote{\url{https://www.wiktionary.org/}}
to extract relations between words that can be encoded on our graph edges.

\section{Conclusion}
We have presented a graph-based semi-supervised method to construct large
annotated morpho-syntactic lexicons from small seed lexicons. Our method is
language independent and we have constructed lexicons for 11 different
languages. We showed that the lexicons thus constructed help improve
performance in morphological tagging and dependency parsing, when used as
features.

\section*{Acknowledgement}
This work was performed when the first author was an intern at Google.
We thank action editor Alexander Clark, and the three anonymous reviewers
for their helpful suggestions in preparing the manuscript.
We thank David Talbot for his help in developing the propagation framework and
helpful discussions about evaluation. We thank Avneesh Saluja, Chris Dyer and
Partha Pratim Talukdar for their comments on drafts of this paper.

\bibliographystyle{acl2012}
\bibliography{references}

\begin{thebibliography}{}

\bibitem[\protect\citename{Agi{\'c} \bgroup et al.\egroup }2015]{agic:2015}
{\v Z}eljko Agi{\'c}, Maria~Jesus Aranzabe, Aitziber Atutxa, Cristina Bosco,
  Jinho Choi, Marie-Catherine de~Marneffe, Timothy Dozat, Rich{\'a}rd Farkas,
  Jennifer Foster, Filip Ginter, Iakes Goenaga, Koldo Gojenola, Yoav Goldberg,
  Jan Haji{\v c}, Anders~Tr{\ae}rup Johannsen, Jenna Kanerva, Juha Kuokkala,
  Veronika Laippala, Alessandro Lenci, Krister Lind{\'e}n, Nikola Ljube{\v
  s}i{\'c}, Teresa Lynn, Christopher Manning, H{\'e}ctor~Alonso Mart{\'i}nez,
  Ryan {McDonald}, Anna Missil{\"a}, Simonetta Montemagni, Joakim Nivre, Hanna
  Nurmi, Petya Osenova, Slav Petrov, Jussi Piitulainen, Barbara Plank, Prokopis
  Prokopidis, Sampo Pyysalo, Wolfgang Seeker, Mojgan Seraji, Natalia Silveira,
  Maria Simi, Kiril Simov, Aaron Smith, Reut Tsarfaty, Veronika Vincze, and
  Daniel Zeman.
\newblock 2015.
\newblock Universal dependencies 1.1.
\newblock {LINDAT}/{CLARIN} digital library at Institute of Formal and Applied
  Linguistics, Charles University in Prague.

\bibitem[\protect\citename{Ahlberg \bgroup et al.\egroup }2014]{ahlberg:14}
Malin Ahlberg, Markus Forsberg, and Mans Hulden.
\newblock 2014.
\newblock Semi-supervised learning of morphological paradigms and lexicons.
\newblock In {\em Proc. of EACL}.

\bibitem[\protect\citename{Ahlberg \bgroup et al.\egroup }2015]{ahlberg:15}
Malin Ahlberg, Markus Forsberg, and Mans Hulden.
\newblock 2015.
\newblock Paradigm classification in supervised learning of morphology.
\newblock In {\em Proc. of NAACL}.

\bibitem[\protect\citename{Alfonseca \bgroup et al.\egroup
  }2010]{alfonseca:2010}
Enrique Alfonseca, Marius Pasca, and Enrique Robledo-Arnuncio.
\newblock 2010.
\newblock Acquisition of instance attributes via labeled and related instances.
\newblock In {\em Proc. of SIGIR}.

\bibitem[\protect\citename{Arisoy \bgroup et al.\egroup
  }2010]{arisoy2010syntactic}
Ebru Arisoy, Murat Sara{\c{c}}lar, Brian Roark, and Izhak Shafran.
\newblock 2010.
\newblock Syntactic and sub-lexical features for {T}urkish discriminative
  language models.
\newblock In {\em Proc. of ICASSP}.

\bibitem[\protect\citename{Banea \bgroup et al.\egroup }2008]{banea2008}
Carmen Banea, Janyce~M. Wiebe, and Rada Mihalcea.
\newblock 2008.
\newblock A bootstrapping method for building subjectivity lexicons for
  languages with scarce resources.
\newblock In {\em Proc. of LREC}.

\bibitem[\protect\citename{Bengio \bgroup et al.\egroup }2006]{Bengio:2006}
Yoshua Bengio, Olivier Delalleau, and Nicolas {Le Roux}.
\newblock 2006.
\newblock Label propagation and quadratic criterion.
\newblock In {\em Semi-Supervised Learning}. MIT Press.

\bibitem[\protect\citename{Besacier \bgroup et al.\egroup
  }2014]{besacier2014automatic}
Laurent Besacier, Etienne Barnard, Alexey Karpov, and Tanja Schultz.
\newblock 2014.
\newblock Automatic speech recognition for under-resourced languages: A survey.
\newblock {\em Speech Communication}, 56:85--100.

\bibitem[\protect\citename{Bohnet and Nivre}2012]{bohnet2012}
Bernd Bohnet and Joakim Nivre.
\newblock 2012.
\newblock A transition-based system for joint part-of-speech tagging and
  labeled non-projective dependency parsing.
\newblock In {\em Proc. of EMNLP}.

\bibitem[\protect\citename{Callison-Burch and Dredze}2010]{Callison-Burch:2010}
Chris Callison-Burch and Mark Dredze.
\newblock 2010.
\newblock Creating speech and language data with amazon's mechanical turk.
\newblock In {\em Proc. of NAACL Workshop on Creating Speech and Language Data
  with {A}mazon's {M}echanical {T}urk}.

\bibitem[\protect\citename{Clark}2003]{clark:2003}
Alexander Clark.
\newblock 2003.
\newblock Combining distributional and morphological information for part of
  speech induction.
\newblock In {\em Proc. of EACL}.

\bibitem[\protect\citename{Creutz and Lagus}2007]{creutz2005unsupervised}
Mathias Creutz and Krista Lagus.
\newblock 2007.
\newblock Unsupervised models for morpheme segmentation and morphology
  learning.
\newblock {\em ACM Transactions on Speech and Language Processing (TSLP)},
  4(1):3.

\bibitem[\protect\citename{Das and Petrov}2011]{das-petrov:2011:ACL-HLT2011}
Dipanjan Das and Slav Petrov.
\newblock 2011.
\newblock Unsupervised part-of-speech tagging with bilingual graph-based
  projections.
\newblock In {\em Proc. of ACL}.

\bibitem[\protect\citename{Das and Smith}2012]{das-smith12}
Dipanjan Das and Noah~A. Smith.
\newblock 2012.
\newblock Graph-based lexicon expansion with sparsity-inducing penalties.
\newblock In {\em Proc. of NAACL}.

\bibitem[\protect\citename{De~Marneffe \bgroup et al.\egroup
  }2014]{de2014universal}
Marie-Catherine De~Marneffe, Timothy Dozat, Natalia Silveira, Katri Haverinen,
  Filip Ginter, Joakim Nivre, and Christopher~D. Manning.
\newblock 2014.
\newblock Universal stanford dependencies: A cross-linguistic typology.
\newblock In {\em Proceedings of LREC}.

\bibitem[\protect\citename{Demberg}2007]{demberg2007language}
Vera Demberg.
\newblock 2007.
\newblock A language-independent unsupervised model for morphological
  segmentation.
\newblock In {\em Proc. of ACL}.

\bibitem[\protect\citename{Denis and Sagot}2009]{denis2009coupling}
Pascal Denis and Beno{\^\i}t Sagot.
\newblock 2009.
\newblock Coupling an annotated corpus and a morphosyntactic lexicon for
  state-of-the-art {POS} tagging with less human effort.
\newblock In {\em Proc. of PACLIC}.

\bibitem[\protect\citename{Dhillon \bgroup et al.\egroup }2012]{dhillon:2012}
Paramveer~S. Dhillon, Partha Talukdar, and Koby Crammer.
\newblock 2012.
\newblock Metric learning for graph-based domain adaptation.
\newblock In {\em Proc. of COLING}.

\bibitem[\protect\citename{Dreyer and Eisner}2011]{de:11}
Markus Dreyer and Jason Eisner.
\newblock 2011.
\newblock Discovering morphological paradigms from plain text using a
  {D}irichlet process mixture model.
\newblock In {\em Proc. of EMNLP}.

\bibitem[\protect\citename{Dryer}2013]{wals-26}
Matthew~S. Dryer.
\newblock 2013.
\newblock {\em Prefixing vs. Suffixing in Inflectional Morphology}.
\newblock Max Planck Institute for Evolutionary Anthropology.

\bibitem[\protect\citename{Duchi \bgroup et al.\egroup
  }2011]{duchi2011adaptive}
John Duchi, Elad Hazan, and Yoram Singer.
\newblock 2011.
\newblock Adaptive subgradient methods for online learning and stochastic
  optimization.
\newblock {\em The Journal of Machine Learning Research}, 12:2121--2159.

\bibitem[\protect\citename{Dukes and Habash}2010]{dukes:2010}
Kais Dukes and Nizar Habash.
\newblock 2010.
\newblock Morphological annotation of quranic {A}rabic.
\newblock In {\em Proc. of LREC}.

\bibitem[\protect\citename{Durrett and DeNero}2013]{ddn:13}
Greg Durrett and John DeNero.
\newblock 2013.
\newblock Supervised learning of complete morphological paradigms.
\newblock In {\em Proc. of NAACL}.

\bibitem[\protect\citename{Dzeroski \bgroup et al.\egroup }2000]{dzeroski:00}
Saso Dzeroski, Tomaz Erjavec, and Jakub Zavrel.
\newblock 2000.
\newblock Morphosyntactic tagging of {S}lovene: Evaluating taggers and tagsets.
\newblock In {\em Proc. of LREC}.

\bibitem[\protect\citename{Erjavec}2004]{erjavec2004multext}
Tomaz Erjavec.
\newblock 2004.
\newblock Multext-east version 3: Multilingual morphosyntactic specifications,
  lexicons and corpora.
\newblock In {\em Proc. of LREC}.

\bibitem[\protect\citename{Erkan and Radev}2004]{erkan2004lexrank}
G\"{u}nes Erkan and Dragomir~R. Radev.
\newblock 2004.
\newblock Lexrank: Graph-based lexical centrality as salience in text
  summarization.
\newblock {\em Journal of Artificial Intelligence Research}, 22(1):457--479.

\bibitem[\protect\citename{Eskander \bgroup et al.\egroup }2013]{eskander:2013}
Ramy Eskander, Nizar Habash, and Owen Rambow.
\newblock 2013.
\newblock Automatic extraction of morphological lexicons from morphologically
  annotated corpora.
\newblock In {\em Proc. of EMNLP}.

\bibitem[\protect\citename{Faruqui and Pad{\'o}}2010]{faruqui:2010}
Manaal Faruqui and Sebastian Pad{\'o}.
\newblock 2010.
\newblock Training and evaluating a {G}erman named entity recognizer with
  semantic generalization.
\newblock In {\em Proc. of KONVENS}.

\bibitem[\protect\citename{Faruqui \bgroup et al.\egroup
  }2016]{faruqui:2016:infl}
Manaal Faruqui, Yulia Tsvetkov, Graham Neubig, and Chris Dyer.
\newblock 2016.
\newblock Morphological inflection generation using character sequence to
  sequence learning.
\newblock {\em arXiv:1104.2086}.

\bibitem[\protect\citename{Ferng and Lin}2011]{ferng2011multi}
Chun-Sung Ferng and Hsuan-Tien Lin.
\newblock 2011.
\newblock Multi-label classification with error-correcting codes.
\newblock In {\em Proc. of ACML}.

\bibitem[\protect\citename{F{\"u}rnkranz \bgroup et al.\egroup
  }2008]{furnkranz:2008}
Johannes F{\"u}rnkranz, Eyke H{\"u}llermeier, Eneldo~Loza Menc{\'\i}a, and
  Klaus Brinker.
\newblock 2008.
\newblock Multilabel classification via calibrated label ranking.
\newblock {\em Machine learning}, 73(2):133--153.

\bibitem[\protect\citename{Garrette \bgroup et al.\egroup
  }2013]{garrette2013real}
Dan Garrette, Jason Mielens, and Jason Baldridge.
\newblock 2013.
\newblock Real-world semi-supervised learning of {POS}-taggers for low-resource
  languages.
\newblock In {\em Proc. of ACL}.

\bibitem[\protect\citename{Ghamrawi and McCallum}2005]{ghamrawi:2005}
Nadia Ghamrawi and Andrew McCallum.
\newblock 2005.
\newblock Collective multi-label classification.
\newblock In {\em Proc. of CIKM}.

\bibitem[\protect\citename{Gim{\'e}nez and Marquez}2004]{marquez2004}
Jes{\'u}s Gim{\'e}nez and Llu{\'i}s Marquez.
\newblock 2004.
\newblock Svmtool: A general {POS} tagger generator based on support vector
  machines.
\newblock In {\em Proc. of LREC}.

\bibitem[\protect\citename{Godbole and Sarawagi}2004]{godbole:2004}
Shantanu Godbole and Sunita Sarawagi.
\newblock 2004.
\newblock Discriminative methods for multi-labeled classification.
\newblock In {\em Proc. of KDD}.

\bibitem[\protect\citename{Goldberg \bgroup et al.\egroup
  }2007]{goldberg2007dissimilarity}
Andrew~B. Goldberg, Xiaojin Zhu, and Stephen~J. Wright.
\newblock 2007.
\newblock Dissimilarity in graph-based semi-supervised classification.
\newblock In {\em Proc. of AISTATS}.

\bibitem[\protect\citename{Goldberg \bgroup et al.\egroup }2009]{Goldberg:2009}
Yoav Goldberg, Reut Tsarfaty, Meni Adler, and Michael Elhadad.
\newblock 2009.
\newblock Enhancing unlexicalized parsing performance using a wide coverage
  lexicon, fuzzy tag-set mapping, and {EM-HMM}-based lexical probabilities.
\newblock In {\em Proc. of EACL}.

\bibitem[\protect\citename{Green and DeNero}2012]{Green:2012}
Spence Green and John DeNero.
\newblock 2012.
\newblock A class-based agreement model for generating accurately inflected
  translations.
\newblock In {\em Proc. of ACL}.

\bibitem[\protect\citename{Haji{\v{c}} and Hladk{\'a}}1998]{hajivc1998}
Jan Haji{\v{c}} and Barbora Hladk{\'a}.
\newblock 1998.
\newblock Tagging inflective languages: Prediction of morphological categories
  for a rich, structured tagset.
\newblock In {\em Proc. of COLING}.

\bibitem[\protect\citename{Haji{\v{c}}}2000]{hajivc2000}
Jan Haji{\v{c}}.
\newblock 2000.
\newblock Morphological tagging: Data vs. dictionaries.
\newblock In {\em Proc. of NAACL}.

\bibitem[\protect\citename{Hsu \bgroup et al.\egroup }2009]{hsu2009multi}
Daniel Hsu, Sham Kakade, John Langford, and Tong Zhang.
\newblock 2009.
\newblock Multi-label prediction via compressed sensing.
\newblock In {\em Proc. of NIPS}.

\bibitem[\protect\citename{Huet \bgroup et al.\egroup }2008]{huet2008}
St{\'e}phane Huet, Guillaume Gravier, and Pascale S{\'e}billot.
\newblock 2008.
\newblock Morphosyntactic resources for automatic speech recognition.
\newblock In {\em Proc. of LREC}.

\bibitem[\protect\citename{Irvine and Klementiev}2010]{Irvine:201}
Ann Irvine and Alexandre Klementiev.
\newblock 2010.
\newblock Using mechanical turk to annotate lexicons for less commonly used
  languages.
\newblock In {\em Proc. of NAACL Workshop on Creating Speech and Language Data
  with {A}mazon's {M}echanical {T}urk}.

\bibitem[\protect\citename{Ising}1925]{ising1925beitrag}
Ernst Ising.
\newblock 1925.
\newblock Beitrag zur theorie des ferromagnetismus.
\newblock {\em Zeitschrift f{\"u}r Physik A Hadrons and Nuclei},
  31(1):253--258.

\bibitem[\protect\citename{Kneser and Ney}1993]{kneser1993improved}
Reinhard Kneser and Hermann Ney.
\newblock 1993.
\newblock Improved clustering techniques for class-based statistical language
  modelling.
\newblock In {\em Proc. of Eurospeech}.

\bibitem[\protect\citename{Kokkinakis \bgroup et al.\egroup
  }2000]{kokkinakis:2000}
Dimitrios Kokkinakis, Maria Toporowska-Gronostaj, and Karin Warmenius.
\newblock 2000.
\newblock Annotating, disambiguating \& automatically extending the coverage of
  the {S}wedish {SIMPLE} lexicon.
\newblock In {\em Proc. of LREC}.

\bibitem[\protect\citename{Koo \bgroup et al.\egroup }2008]{koo:2008}
Terry Koo, Xavier Carreras, and Michael Collins.
\newblock 2008.
\newblock Simple semi-supervised dependency parsing.
\newblock In {\em Proc. of ACL}.

\bibitem[\protect\citename{K{\"u}bler \bgroup et al.\egroup }2009]{kubler2009}
Sandra K{\"u}bler, Ryan McDonald, and Joakim Nivre.
\newblock 2009.
\newblock {\em Dependency parsing}.
\newblock Synthesis Lectures on Human Language Technologies. Morgan \& Claypool
  Publishers.

\bibitem[\protect\citename{Lu \bgroup et al.\egroup }2011]{Lu:2011}
Yue Lu, Malu Castellanos, Umeshwar Dayal, and ChengXiang Zhai.
\newblock 2011.
\newblock Automatic construction of a context-aware sentiment lexicon: An
  optimization approach.
\newblock In {\em Proc. of WWW}.

\bibitem[\protect\citename{Martin \bgroup et al.\egroup
  }1998]{martin1998algorithms}
Sven Martin, J{\"o}rg Liermann, and Hermann Ney.
\newblock 1998.
\newblock Algorithms for bigram and trigram word clustering.
\newblock {\em Speech communication}, 24(1):19--37.

\bibitem[\protect\citename{McDonald \bgroup et al.\egroup }2013]{mcdonald:2013}
Ryan McDonald, Joakim Nivre, Yvonne Quirmbach-Brundage, Yoav Goldberg, Dipanjan
  Das, Kuzman Ganchev, Keith~B. Hall, Slav Petrov, Hao Zhang, Oscar
  T{\"a}ckstr{\"o}m, Claudia Bedini, N{\'u}ria~B. Castell{\'o}, and Jungmee
  Lee.
\newblock 2013.
\newblock Universal dependency annotation for multilingual parsing.
\newblock In {\em Proc. of ACL}.

\bibitem[\protect\citename{Mikolov \bgroup et al.\egroup
  }2013]{mikolov-yih-zweig:2013:NAACL}
Tomas Mikolov, Wen-tau Yih, and Geoffrey Zweig.
\newblock 2013.
\newblock Linguistic regularities in continuous space word representations.
\newblock In {\em Proc. of NAACL}.

\bibitem[\protect\citename{Minkov \bgroup et al.\egroup
  }2007]{minkov2007generating}
Einat Minkov, Kristina Toutanova, and Hisami Suzuki.
\newblock 2007.
\newblock Generating complex morphology for machine translation.
\newblock In {\em Proc. of ACL}.

\bibitem[\protect\citename{Mohammad \bgroup et al.\egroup }2009]{Mohammad:2009}
Saif Mohammad, Cody Dunne, and Bonnie Dorr.
\newblock 2009.
\newblock Generating high-coverage semantic orientation lexicons from overtly
  marked words and a thesaurus.
\newblock In {\em Proc. of EMNLP}.

\bibitem[\protect\citename{Moore}2015]{moore:2015:EMNLP}
Robert Moore.
\newblock 2015.
\newblock An improved tag dictionary for faster part-of-speech tagging.
\newblock In {\em Proc. of EMNLP}.

\bibitem[\protect\citename{M\"{u}ller and Schuetze}2015]{muller:2015}
Thomas M\"{u}ller and Hinrich Schuetze.
\newblock 2015.
\newblock Robust morphological tagging with word representations.
\newblock In {\em Proceedings of NAACL}.

\bibitem[\protect\citename{M{\"u}ller \bgroup et al.\egroup
  }2013]{muller2013efficient}
Thomas M{\"u}ller, Helmut Schmid, and Hinrich Sch{\"u}tze.
\newblock 2013.
\newblock Efficient higher-order {CRF}s for morphological tagging.
\newblock In {\em Proc. of EMNLP}.

\bibitem[\protect\citename{Muthukrishnan \bgroup et al.\egroup
  }2011]{muthu:2011}
Pradeep Muthukrishnan, Dragomir Radev, and Qiaozhu Mei.
\newblock 2011.
\newblock Simultaneous similarity learning and feature-weight learning for
  document clustering.
\newblock In {\em Proc. of TextGraphs}.

\bibitem[\protect\citename{Narasimhan \bgroup et al.\egroup
  }2015]{narasimhan2015unsupervised}
Karthik Narasimhan, Regina Barzilay, and Tommi Jaakkola.
\newblock 2015.
\newblock An unsupervised method for uncovering morphological chains.
\newblock {\em Transactions of the Association for Computational Linguistics},
  3:157--167.

\bibitem[\protect\citename{Nicolai \bgroup et al.\egroup
  }2015]{kondrak:15:inflection}
Garrett Nicolai, Colin Cherry, and Grzegorz Kondrak.
\newblock 2015.
\newblock Inflection generation as discriminative string transduction.
\newblock In {\em Proc. of NAACL}.

\bibitem[\protect\citename{Nie{\ss}en and Ney}2004]{niessen2004statistical}
Sonja Nie{\ss}en and Hermann Ney.
\newblock 2004.
\newblock Statistical machine translation with scarce resources using
  morpho-syntactic information.
\newblock {\em Computational {L}inguistics}, 30(2).

\bibitem[\protect\citename{Oflazer and Kuru{\"o}z}1994]{oflazer1994tagging}
Kemal Oflazer and {\`I}lker Kuru{\"o}z.
\newblock 1994.
\newblock Tagging and morphological disambiguation of {T}urkish text.
\newblock In {\em Proc. of ANLP}.

\bibitem[\protect\citename{Owoputi \bgroup et al.\egroup }2013]{owoputi:2013}
Olutobi Owoputi, Brendan O'Connor, Chris Dyer, Kevin Gimpel, Nathan Schneider,
  and Noah~A. Smith.
\newblock 2013.
\newblock Improved part-of-speech tagging for online conversational text with
  word clusters.
\newblock In {\em Proc. of NAACL}.

\bibitem[\protect\citename{Pirinen}2011]{pirinen2011}
Tommi~A Pirinen.
\newblock 2011.
\newblock Modularisation of finnish finite-state language description—towards
  wide collaboration in open source development of a morphological analyser.
\newblock In {\em Proc. of {NODALIDA}}.

\bibitem[\protect\citename{Poon \bgroup et al.\egroup
  }2009]{poon2009unsupervised}
Hoifung Poon, Colin Cherry, and Kristina Toutanova.
\newblock 2009.
\newblock Unsupervised morphological segmentation with log-linear models.
\newblock In {\em Proc. of NAACL}.

\bibitem[\protect\citename{Ratnaparkhi}1996]{ratnaparkhi1996maximum}
Adwait Ratnaparkhi.
\newblock 1996.
\newblock A maximum entropy model for part-of-speech tagging.
\newblock In {\em Proc. of EMNLP}.

\bibitem[\protect\citename{Read \bgroup et al.\egroup }2011]{read:2011}
Jesse Read, Bernhard Pfahringer, Geoff Holmes, and Eibe Frank.
\newblock 2011.
\newblock Classifier chains for multi-label classification.
\newblock {\em Machine Learning}, 85(3):333--359.

\bibitem[\protect\citename{Sak \bgroup et al.\egroup }2008]{sak2008turkish}
Ha{\c{s}}im Sak, Tunga G{\"u}ng{\"o}r, and Murat Sara{\c{c}}lar.
\newblock 2008.
\newblock {T}urkish language resources: Morphological parser, morphological
  disambiguator and web corpus.
\newblock In {\em Proc. of ANLP}.

\bibitem[\protect\citename{Saluja and Navr{\'a}til}2013]{saluja:2013}
Avneesh Saluja and Jir{\i} Navr{\'a}til.
\newblock 2013.
\newblock Graph-based unsupervised learning of word similarities using
  heterogeneous feature types.
\newblock In {\em Proc. of TextGraphs}.

\bibitem[\protect\citename{Schmid}1994]{schmid1994}
Helmut Schmid.
\newblock 1994.
\newblock Probabilistic part-of-speech tagging using decision trees.
\newblock In {\em Proc. of the International Conference on New Methods in
  Language Processing}.

\bibitem[\protect\citename{Simov \bgroup et al.\egroup
  }2004]{simov2004language}
Kiril~Ivanov Simov, Petya Osenova, Sia Kolkovska, Elisaveta Balabanova, and
  Dimitar Doikoff.
\newblock 2004.
\newblock A language resources infrastructure for {B}ulgarian.
\newblock In {\em Proc. of LREC}.

\bibitem[\protect\citename{Smith \bgroup et al.\egroup }2005]{smith2005context}
Noah~A. Smith, David~A. Smith, and Roy~W. Tromble.
\newblock 2005.
\newblock Context-based morphological disambiguation with random fields.
\newblock In {\em Proc. of EMNLP}.

\bibitem[\protect\citename{Snyder and Barzilay}2008]{snyder2008unsupervised}
Benjamin Snyder and Regina Barzilay.
\newblock 2008.
\newblock Unsupervised multilingual learning for morphological segmentation.
\newblock In {\em Proc. of ACL}.

\bibitem[\protect\citename{Soricut and Och}2015]{soricut:2015}
Radu Soricut and Franz Och.
\newblock 2015.
\newblock Unsupervised morphology induction using word embeddings.
\newblock In {\em Proc. of NAACL}.

\bibitem[\protect\citename{Subramanya and Bilmes}2008]{subramanya2008soft}
Amarnag Subramanya and Jeff Bilmes.
\newblock 2008.
\newblock Soft-supervised learning for text classification.
\newblock In {\em Proc. of EMNLP}.

\bibitem[\protect\citename{Subramanya and Talukdar}2014]{subramanya2014graph}
Amarnag Subramanya and Partha~Pratim Talukdar.
\newblock 2014.
\newblock Graph-based semi-supervised learning.
\newblock {\em Synthesis Lectures on Artificial Intelligence and Machine
  Learning}, 8(4).

\bibitem[\protect\citename{Subramanya \bgroup et al.\egroup
  }2010]{Subramanya:2010}
Amarnag Subramanya, Slav Petrov, and Fernando Pereira.
\newblock 2010.
\newblock Efficient graph-based semi-supervised learning of structured tagging
  models.
\newblock In {\em Proc. of EMNLP}.

\bibitem[\protect\citename{T{\"a}ckstr{\"o}m \bgroup et al.\egroup
  }2012]{tackstrom:2012}
Oscar T{\"a}ckstr{\"o}m, Ryan McDonald, and Jakob Uszkoreit.
\newblock 2012.
\newblock Cross-lingual word clusters for direct transfer of linguistic
  structure.
\newblock In {\em Proc. of NAACL}.

\bibitem[\protect\citename{Takamura \bgroup et al.\egroup }2007]{takamura}
Hiroya Takamura, Takashi Inui, and Manabu Okumura.
\newblock 2007.
\newblock Extracting semantic orientations of phrases from dictionary.
\newblock In {\em Proc. of NAACL}.

\bibitem[\protect\citename{Talukdar and Cohen}2013]{talukdar:2013}
Partha~Pratim Talukdar and William Cohen.
\newblock 2013.
\newblock Scaling graph-based semi-supervised learning to large number of
  labels using count-min sketch.
\newblock In {\em Proc. of AISTATS}.

\bibitem[\protect\citename{Talukdar and Pereira}2010]{talukdar2010experiments}
Partha~Pratim Talukdar and Fernando Pereira.
\newblock 2010.
\newblock Experiments in graph-based semi-supervised learning methods for
  class-instance acquisition.
\newblock In {\em Proc. of ACL}.

\bibitem[\protect\citename{Talukdar \bgroup et al.\egroup
  }2012]{talukdar2012acquiring}
Partha~Pratim Talukdar, Derry Wijaya, and Tom Mitchell.
\newblock 2012.
\newblock Acquiring temporal constraints between relations.
\newblock In {\em Proc. of CIKM}.

\bibitem[\protect\citename{Thelen and Riloff}2002]{Thelen:2002}
Michael Thelen and Ellen Riloff.
\newblock 2002.
\newblock A bootstrapping method for learning semantic lexicons using
  extraction pattern contexts.
\newblock In {\em Proc. of ACL}.

\bibitem[\protect\citename{Tr{\'o}n \bgroup et al.\egroup }2006]{tron2006}
Viktor Tr{\'o}n, P{\'e}ter Hal{\'a}csy, P{\'e}ter Rebrus, Andr{\'a}s Rung,
  P{\'e}ter Vajda, and Eszter Simon.
\newblock 2006.
\newblock Morphdb.hu: Hungarian lexical database and morphological grammar.
\newblock In {\em Proc. of LREC}.

\bibitem[\protect\citename{Tsoumakas and Katakis}2006]{tsoumakas:2006}
Grigorios Tsoumakas and Ioannis Katakis.
\newblock 2006.
\newblock Multi-label classification: An overview.
\newblock {\em Dept. of Informatics, Aristotle University of Thessaloniki,
  Greece}.

\bibitem[\protect\citename{Turian \bgroup et al.\egroup }2010]{turian:2010}
Joseph Turian, Lev Ratinov, and Yoshua Bengio.
\newblock 2010.
\newblock Word representations: a simple and general method for semi-supervised
  learning.
\newblock In {\em Proc. of ACL}.

\bibitem[\protect\citename{Uszkoreit and Brants}2008]{uszkoreit2008distributed}
Jakob Uszkoreit and Thorsten Brants.
\newblock 2008.
\newblock Distributed word clustering for large scale class-based language
  modeling in machine translation.
\newblock In {\em Proc. of ACL}.

\bibitem[\protect\citename{Velikovich \bgroup et al.\egroup
  }2010]{velikovich2010}
Leonid Velikovich, Sasha Blair-Goldensohn, Kerry Hannan, and Ryan McDonald.
\newblock 2010.
\newblock The viability of web-derived polarity lexicons.
\newblock In {\em Proc. of NAACL}.

\bibitem[\protect\citename{Wang \bgroup et al.\egroup }2007]{wang2007semi}
Fei Wang, Shijun Wang, Changshui Zhang, and Ole Winther.
\newblock 2007.
\newblock Semi-supervised mean fields.
\newblock In {\em Proc. of AISTATS}.

\bibitem[\protect\citename{Weinberger \bgroup et al.\egroup
  }2005]{weinberger2005}
Kilian~Q. Weinberger, John Blitzer, and Lawrence~K. Saul.
\newblock 2005.
\newblock Distance metric learning for large margin nearest neighbor
  classification.
\newblock In {\em Proc. of NIPS}.

\bibitem[\protect\citename{Wicentowski}2004]{wicentowski:04}
Richard Wicentowski.
\newblock 2004.
\newblock Multilingual noise-robust supervised morphological analysis using the
  wordframe model.
\newblock In {\em Proc. of SIGPHON}.

\bibitem[\protect\citename{Yarowsky and Wicentowski}2000]{yarowsky:00}
David Yarowsky and Richard Wicentowski.
\newblock 2000.
\newblock Minimally supervised morphological analysis by multimodal alignment.
\newblock In {\em Proc. of ACL}.

\bibitem[\protect\citename{Zhang and Nivre}2011]{zhang2011}
Yue Zhang and Joakim Nivre.
\newblock 2011.
\newblock Transition-based dependency parsing with rich non-local features.
\newblock In {\em Proc. of ACL}.

\bibitem[\protect\citename{Zhang and Schneider}2011]{zhang2011multi}
Yi~Zhang and Jeff~G. Schneider.
\newblock 2011.
\newblock Multi-label output codes using canonical correlation analysis.
\newblock In {\em Proc. of AISTATS}.

\bibitem[\protect\citename{Zhang and Zhou}2005]{zhang:2005}
Min-Ling Zhang and Zhi-Hua Zhou.
\newblock 2005.
\newblock A k-nearest neighbor based algorithm for multi-label classification.
\newblock In {\em Proc. of IEEE Conference on Granular Computing}.

\bibitem[\protect\citename{Zhu}2005]{zhu2005semi}
Xiaojin Zhu.
\newblock 2005.
\newblock {\em Semi-supervised Learning with Graphs}.
\newblock {Ph.D.} thesis, Carnegie Mellon University, Pittsburgh, PA, USA.
\newblock AAI3179046.

\end{thebibliography}
\end{document}